\documentclass[letterpaper]{article} 
\usepackage{aaai24}  
\usepackage{times}  
\usepackage{helvet}  
\usepackage{courier}  
\usepackage[hyphens]{url}  
\usepackage{graphicx} 
\usepackage{natbib}  
\usepackage{caption} 
\usepackage{algorithm}
\usepackage{algorithmic}
\usepackage{newfloat}
\usepackage{listings}
\DeclareCaptionStyle{ruled}{labelfont=normalfont,labelsep=colon,strut=off} 
\floatstyle{ruled}
\newfloat{listing}{tb}{lst}{}
\floatname{listing}{Listing}
\usepackage{epsfig}
\usepackage{amsmath}
\usepackage{tabularx}
\usepackage{booktabs}
\usepackage{pifont}
\usepackage{multirow}

\newcommand{\data}{\textsc{CSTBIR}}
\newcommand{\model}{\textsc{STNet}}

\newcommand{\loss}[1]{$\mathcal{L}_{#1}$}
\title{Composite Sketch+Text Queries for Retrieving Objects with \\Elusive Names and Complex Interactions}
\author{
    Prajwal Gatti\textsuperscript{\rm 1}, Kshitij Parikh\textsuperscript{\rm 1}, Dhriti Prasanna Paul\textsuperscript{\rm 1}, Manish Gupta\textsuperscript{\rm 2}, Anand Mishra\textsuperscript{\rm 1} \\
}
\affiliations{
    \textsuperscript{\rm 1}Indian Institute of Technology Jodhpur\\
    \textsuperscript{\rm 2}Microsoft\\
      \{pgatti, parikh.2, paul.4, mishra\}@iitj.ac.in, gmanish@microsoft.com 
}

\begin{document}

\maketitle

\begin{abstract}
Non-native speakers with limited vocabulary often struggle to name specific objects despite being able to visualize them, e.g., people outside Australia searching for `numbats.' Further, users may want to search for such elusive objects with difficult-to-sketch interactions, e.g., ``numbat digging in the ground.'' In such common but complex situations, users desire a search interface that accepts composite multimodal queries comprising hand-drawn sketches of ``difficult-to-name but easy-to-draw'' objects and text describing ``difficult-to-sketch but easy-to-verbalize'' object's attributes or interaction with the scene. This novel problem statement distinctly differs from the previously well-researched TBIR (text-based image retrieval) and SBIR (sketch-based image retrieval) problems. To study this under-explored task, we curate a dataset, \data{} (\underline{C}omposite \underline{S}ketch+\underline{T}ext \underline{B}ased \underline{I}mage \underline{R}etrieval), consisting of $\sim$2M queries and 108K natural scene images. Further, as a solution to this problem, we propose a pretrained multimodal transformer-based baseline, \model{} (\underline{S}ketch+\underline{T}ext \underline{Net}work), that uses a hand-drawn sketch to localize relevant objects in the natural scene image, and encodes the text and image to perform image retrieval. In addition to contrastive learning, we propose multiple training objectives that improve the performance of our model. Extensive experiments show that our proposed method outperforms several state-of-the-art retrieval methods for text-only, sketch-only, and composite query modalities. We make the dataset and code available at: \url{https://vl2g.github.io/projects/cstbir}.
\end{abstract}

\section{Introduction}
\label{sec:introduction}

\begin{figure}[!t]
        \centering
    \includegraphics[width=\columnwidth]{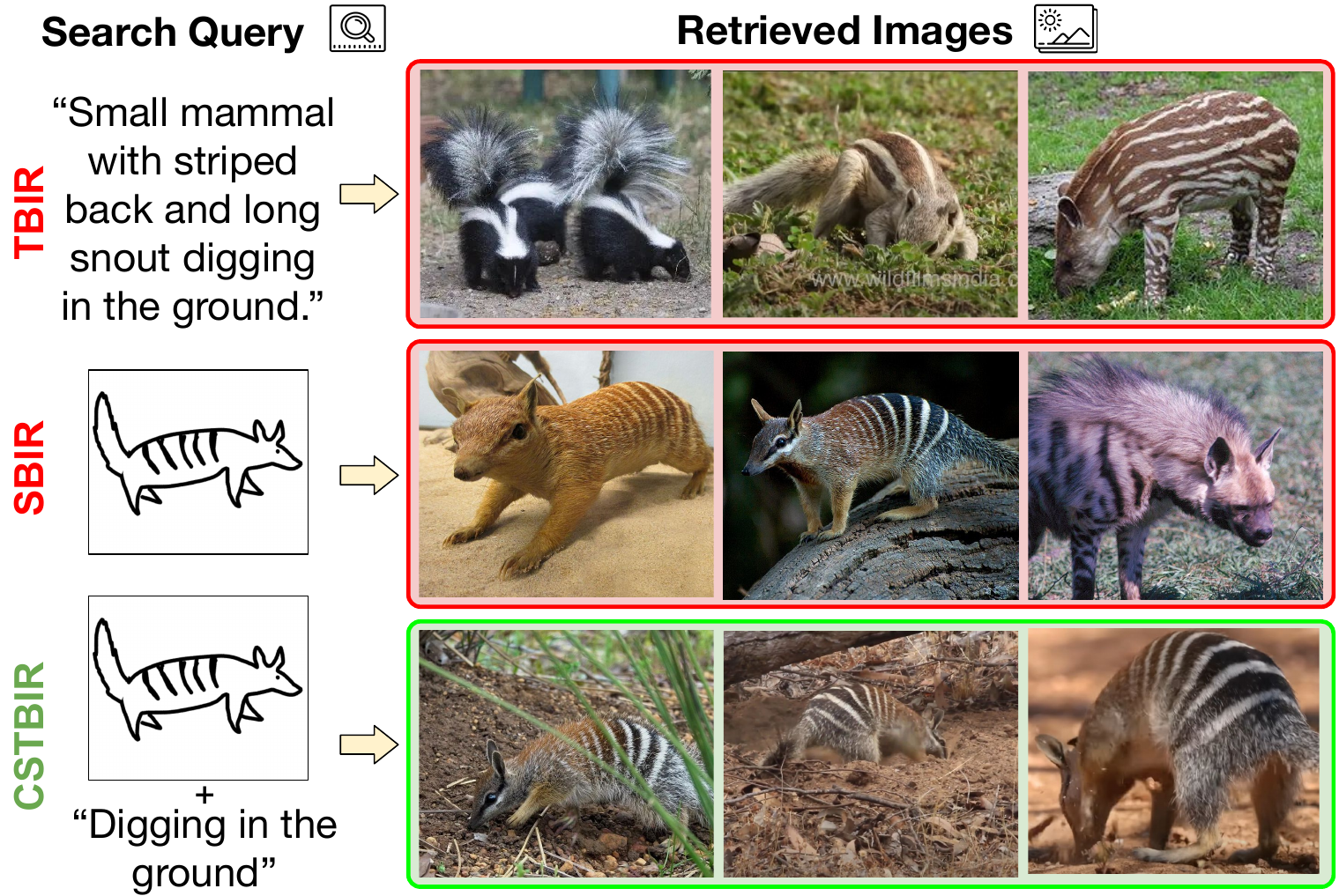}
          \caption{\data{}: Composite Sketch+Text Based Image Retrieval: 
          A user wants to search ``Numbat digging in the ground'' but does not know the word ``numbat'', and the interaction ``digging in the ground'' is not easy to sketch.
          }
          \label{fig:introduction}
    \end{figure}

Traditional text-based image retrieval (TBIR) systems~\cite{li2019visualbert,kim2021vilt,zhang2020context,lee2018stacked,li2020oscar} are intuitive for users with strong linguistic abilities. However, non-native speakers or users unfamiliar with particular objects struggle in using such systems to find objects with ``elusive" names, e.g., users outside Australia searching for numbats, as shown in Figure~\ref{fig:introduction}. Elaborate text descriptions in lieu of the precise object name could provide limited help, even with all the details. For example, ``Small mammal with striped back and long snout digging in the ground" as a replacement for ``numbats'' leads to images of chipmunk, badger, weasel, mongoose, or skunk.

Sketch-based image retrieval (SBIR) systems~\cite{yu2016sketch, dey2019doodle, song2017deep, collomosse2019livesketch, sain2022sketch3t} seem to provide an illusory relief in such situations. Although a user can sketch ``difficult-to-name but easy-to-draw'' objects, (1) users may not have enough time, skills, or tools to draw all the details, leading to ambiguity in sketches; (2) users may be looking for ``difficult-to-sketch but easy-to-verbalize'' object's attributes or interaction with the scene. For example, for the query, ``numbat digging in the ground'', it is difficult to draw a sketch to represent ``digging in the ground'', and even if drawn, the sketch could lead to false positives about ``numbat eating an insect'' or ``numbat searching for termites''. 

Such common but complex search situations require novel multimodal search interfaces, allowing seamless text and sketch mix-ups in queries. Such a flexible and natural interface should help the user to draw sketches for ``difficult-to-name'' objects effortlessly and then complement their creations with text descriptions to define layout, color, pose, and other object characteristics, along with complex interactions with other objects in scenes. We refer to such a novel proposed system as \data{} or \underline{\textbf{C}}omposite \underline{\textbf{S}}ketch+\underline{\textbf{T}}ext \underline{\textbf{B}}ased \underline{\textbf{I}}mage \underline{\textbf{R}}etrieval system.

Although a vast literature exists on TBIR and SBIR, to the best of our knowledge, the \data{} problem setting has yet to be studied rigorously. There have been some recent works~\cite{song2017fine,sangkloy2022sketch,chowdhury2023scenetrilogy} that attempt to solve a simpler version, where: (a) target image collection is focused objects rather than complex natural scenes, (b) sketch is at scene-level rather than object-level, or (c) text description is comprehensive rather than partial (or complementary). This paper proposes a system for the complex \data{} setting.

Given input with a rough sketch and a complementary text description (e.g., the sketch of ``numbat'' and text=``digging in the ground''), an evident approach is to guess the object name from the sketch, complete the text query as ``Numbat digging in the ground'' and then use TBIR methods. However, such a two-stage method may fail when the sketch represents an object with an ambiguous name (e.g., mouse, bat, crane) and suffers from signal loss when attempting to describe knowledge in the sketch using an object name. Additionally, such two-stage approaches are restricted to closed-world settings where the object names are previously known and may not generalize well to rare or novel objects. Hence, we propose a principled method -- \model{} that jointly processes text and sketch inputs. More specifically, we propose the following task-specific pretraining objectives for the multimodal transformer: (i) object classification, i.e., predict object name; (ii) sketch-guided object detection, i.e., localize the relevant objects in the image, and (iii) sketch reconstruction, i.e., recreate the query sketch from the multimodal representation of sketch, and the image.  

Overall, we make the following main contributions in this paper: 
(i) We study an important and under-explored task, namely \data{}. (ii) Toward this novel setting, we contribute a large dataset of $\sim$2M queries and $\sim$108K natural scene images. (iii) For \data{}, we pre-train a multimodal Transformer \model{}, designed to handle sketch and text as inputs, using multiple loss functions: contrastive loss, object classification loss, sketch-guided object detection loss, and sketch reconstruction loss. (iv) Our proposed model outperforms several competitive text-only, sketch-only, and sketch+text baselines. 

\section{Related Work}
Image retrieval systems can answer queries expressed using hand-drawn sketches (SBIR), text (TBIR), a combination of sketch and text (\data{}), color layout, concept layout~\cite{zhou2017recent}, visual features~\cite{tian2023fashion,dodds2020modality}, or location-sensitive tags~\cite{gomez2020location}. We review existing work on TBIR, SBIR, and multimodal query-based IR.

\paragraph{Sketch-Based Image Retrieval (SBIR):}
It allows the flexibility to easily specify the qualitative characteristics using sketches~\cite{yu2016sketch, dey2019doodle, song2017deep}. Following the initial work on sketch recognition~\cite{sun2012sketch2tag}, earlier SBIR studies mainly focused on convolutional neural networks (CNN)~\cite{yu2016sketch, liu2017deep} which was soon followed by various Transformer~\cite{vaswani2017attention}-based architectures~\cite{Ribeiro2020SketchformerTR, chowdhury2022fs}. Deep Siamese models with triplet loss have also been explored~\cite{yu2016sketch, collomosse2019livesketch}. Several specialized SBIR settings have also emerged such as Zero Shot-SBIR~\cite{pandey2020stacked,dey2019doodle,dutta2019semantically}, fine-grained SBIR~\cite{liu2020scenesketcher, bhunia2022adaptive, pang2019generalising, pang2017cross,ling2022conditional,bhunia2020sketch,song2017deep}, and category-level SBIR~\cite{sain2021stylemeup,bhunia2021vectorization,sain2022sketch3t}.

\paragraph{Text-Based Image Retrieval (TBIR):}
Popular methods for TBIR include alignment of input text and the corresponding input image using pretrained multimodal Transformer methods like VisualBERT~\cite{li2019visualbert} and ViLT~\cite{kim2021vilt}. Further, cross-attention-based models~\cite{zhang2020context,lee2018stacked} and models that use object tags detected in images~\cite{li2020oscar} have also been proposed. Recently, contrastive learning methods~\cite{jia2021scaling}, along with zero-shot learning~\cite{radford2021learning}, have been shown to achieve state-of-the-art results.

\begin{table}[!t]
\small
    \centering
    \begin{tabular}{|l|l|c|c|c|}
    \hline
     Query&Dataset&Sketch&Text&Image\\
      \hline
      \hline
S& TU-Berlin&Object&None&Object\\
S&QMUL-Shoe-V2&Object&None&Object\\
T&COCO&None&Complete&Scene\\
T&Flickr-30K&None&Complete&Scene\\
S+T&FS COCO&Scene&Complete&Scene\\
S+T&CSTBIR (Ours)&Object&Complementary&Scene\\
\hline
    \end{tabular}
    \caption{Comparison of datasets with \data{}. \data{} uniquely requires searching over a database of natural scene images using queries of object sketch and partial complementary natural language sentences. S: Sketch, T: Text.}
    \label{tab:datasetComparison}
\end{table}

\paragraph{Multimodal Query Based Image Retrieval:}
Several systems have been built to consume multimodal input for image retrieval. 
Earlier works used reference images and text as an attribute on a category-level retrieval~\cite{kovashka2012whittlesearch,han2017automatic}. Input text data was more elaborated to provide improved results~\cite{guo2018dialog,vo2019composing}. While such earlier systems used CNNs, more recent systems~\cite{song2023boosting,baldrati2022effective} leverage Transformers. Further, some studies~\cite{changpinyo2021telling,pont2020connecting} explored the setting where the user simultaneously uses both speech and mouse traces as the query. Lastly,~\cite{nakatsuka2023content} search images relevant to input music.

It is not always possible to have an input reference image for image retrieval; instead, a sketch (along with text description) is used, which gives more flexibility. Image retrieval using hand-drawn sketches and textual descriptive data has been under-explored.

Detailed sketch and text input have been used to (a) retrieve e-commerce product images using CNNs and LSTMs~\cite{song2017fine}, and (b) retrieve scene images using CLIP~\cite{sangkloy2022sketch,chowdhury2023scenetrilogy}. However, in several practical scenarios, 
(a) the sketch is object-level, very rough, and not elaborate, and (b) the text is partial (complementary to sketch) and not self-contained. Unfortunately, no previous work exists for such a (complex) practical setting. Our contributed dataset, \data{}, and the proposed method addresses this setting in this paper. Compared to~\cite{sangkloy2022sketch} where sketch covers 100\% area of the image to be retrieved, in our dataset, sketches cover only 36.7\% area of the matching scene image on average.
In our dataset, sketches are less complex than in~\cite{sangkloy2022sketch}, which contain $\sim$2.6x times more sketch pixels compared to our dataset\footnote{For fair comparison in terms of pixels covered by the sketch strokes, we apply thinning to normalize the stroke width for both datasets:~\cite{sangkloy2022sketch} and ours.}. Table~\ref{tab:datasetComparison} shows these comparisons of \data{} with other existing image retrieval datasets~\cite{eitz2012humans,yu2016sketch,lin2014microsoft,young2014image,chowdhury2022fs}.

\begin{figure}[!t]
    \centering
     \includegraphics[width=\columnwidth]{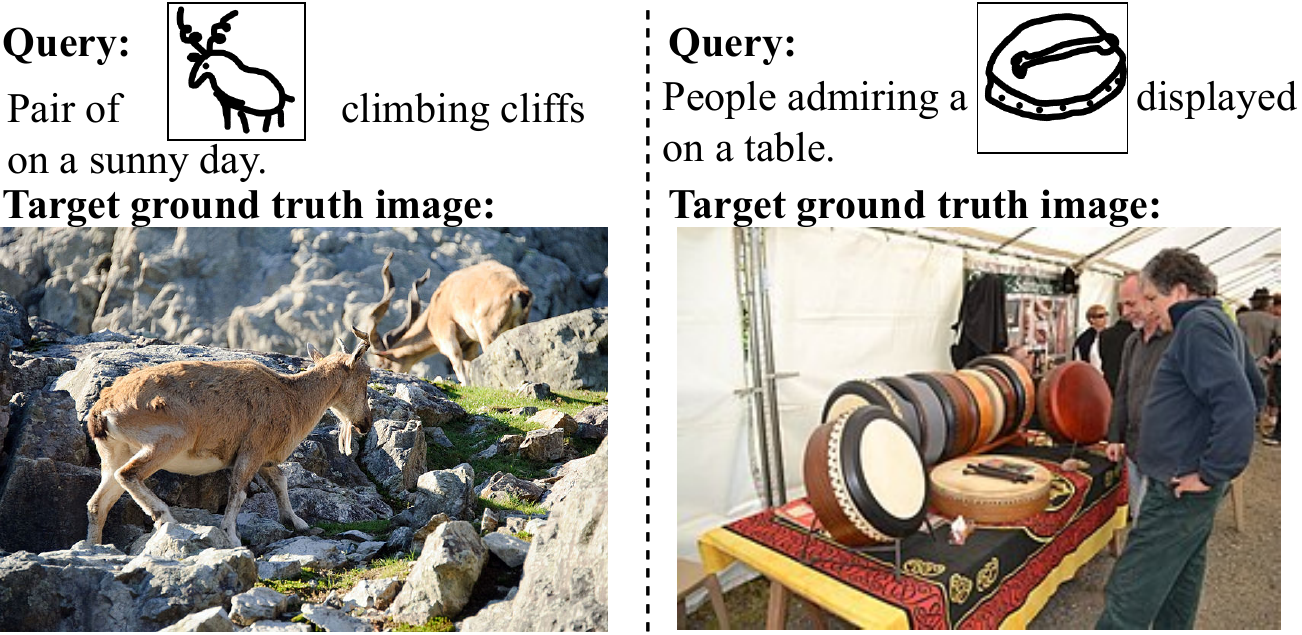}
          \caption{Examples from our dataset -- \data{}. It contains queries composed of a sketch of an object, a natural language text describing its attributes and interactions, and the target natural scene image containing the object. Queried objects: markhor (left), and bodhran (right).}
    \label{fig:datasetExamples}
\end{figure}

\section{The \data{} Problem and Dataset}

Given a hand-drawn sketch $S$, a complementary text $T$ and a database ${\mathcal D} = \{I_{i}\}_{i=1}^{N}$ of natural scene images with multiple objects, the \data{} problem aims to rank the $N$ images according to relevance to the composite $\langle S, T\rangle$ query.

Due to the lack of a suitable dataset, we curate the \data{} dataset, where each sample consists of a hand-drawn sketch, a partial complementary text description, and a relevant natural scene image. The natural scene images in the database have multiple object categories, attributes, relationships, and activities. Although this dataset does not have ``difficult-to-name'' objects, it is a reasonable proxy. We also evaluate using a manually curated separate test set of ``difficult-to-name'' objects.

The natural images and text descriptions are taken from Visual Genome~\cite{krishna2017visual}. The dense annotations in this dataset allow us to frame multiple queries related to an image, each of which pertains to a particular object in the image. The hand-drawn sketches are taken from the Quick, Draw! dataset~\cite{ha2017neural}. Annotators have drawn these sketches in $<$20 seconds; hence, they are rough and lack the exact details as that of an image, which aligns with the challenging real-world setting of this task. Quick, Draw! has over 50M sketches across 345 categories.
We take the intersection of the object categories between Visual Genome and Quick, Draw! to get 258 intersecting object classes in \data{}. This leads to $\sim$108K natural images with $\sim$2M queries in \data{}. We pair each query from Visual Genome with the corresponding object's sketch, sampled randomly from 10K sketches taken for each category from Quick, Draw! 

Table~\ref{tab:dataStats} shows basic statistics of the dataset. The dataset has been split into train, validation, and test based on the corresponding splits from Visual Genome for the scene images. The dataset has a total of $\sim$108K images and $\sim$562K sketches. The training dataset consists of $\sim$97K images, $\sim$484K sketches, and $\sim$1.89M queries. On average, the text sentences contain 5.4 words. The dataset also includes a validation set with $\sim$5K images, $\sim$83K sketches, and $\sim$97K queries. Further, it contains three test sets: Test-1K, Test-5K, and Open-Category set. Test-1K includes 1K queries and corresponding 1K natural scene images in the gallery. Test-5K is a more challenging set that contains 4K queries and 5K gallery images. All the sketch object categories in Test-1K and Test-5K sets are present during training. However, the scene and sketch images in the test set were not part of the training or validation set. We created the Open-Category test set to evaluate the model on novel object categories unseen at train time, which contains 70 novel object categories (of which 50 are ``difficult-to-name'') and corresponding sketches.

Figure~\ref{fig:datasetExamples} shows two examples from the dataset. For further data analysis, we performed part of speech tagging on text descriptions using NLTK. We visualize these statistics in the Appendix as word clouds for the top few adjectives (object attribute indicating words), verbs (action indicating words), and prepositions (position indicating words) for the text descriptions in the \data{} dataset.

\begin{table}[!t]
    \centering
        \small
        \begin{tabular}{|l|c|}
        \hline
        Property&Value\\
        \hline
        \hline
Avg sentence length (in words/tokens) & 5.4 / 7.7 \\
\hline
\# Unique Images & 108K \\
\hline
\# Unique Sketches & 562K \\
\hline
\# Unique Object Categories & 258 \\
\hline
\# Training Instances & 1.89M \\
\hline
\# Validation Instances & 97K\\
\hline
\# Test Instances & 5000 \\
\hline
Avg \% Area Covered by Query & 36.7 \\
\hline
        \end{tabular}
        \captionof{table}{\data{} Dataset Statistics}
        \label{tab:dataStats}
\end{table}

\section{The Proposed \model{} Model for \data{}}
For the task of sketch and text-based image retrieval, we introduce \model{} (\underline{S}ketch+\underline{T}ext \underline{Net}work), a novel multimodal architecture. It comprises three independent Transformer-based encoder networks based on the pretrained CLIP model~\cite{radford2021learning}. The overall architecture of \model{} is illustrated in Figure~\ref{fig: method}. Next, we describe the working and architectural details of \model{} in the following subsections.

\subsection{Query (Sketch+Text) Encoding}
In \data{}, the query consists of a text sentence and a hand-drawn sketch. We independently encode these two inputs using a pretrained CLIP text encoder and a pretrained Vision Transformer (ViT)~\cite{dosovitskiy2020image} encoder. 

Given a query text sentence $T$, we tokenize it using a Byte-Pair-Encoding scheme according to the learned vocabulary of the text encoder as $F_T = [CLS, t_1, t_2, \ldots, t_n]$, where each $t_i$ represents a sub-word token, and $CLS$ is the global pool token. Given the query sketch image $S$, we use a pretrained ViT encoder which is fed the input $F_{S} = [CLS, s_1, s_2, \ldots, s_m]$, where each $s_i$ is the embedding of an image patch. As the ViT encoder is pretrained on the ImageNet-21K dataset~\cite{ridnik2021imagenet21k}, we first train it on the sketch data for the classification task to adapt it for the sketch domain. This trained encoder is then used to embed the sketch input. Overall, this results into text embedding $h_{CLS}^{T}$ and sketch embedding $h_{CLS}^{S}$.

\subsection{Image Encoding}

To utilize the benefits of large-scale pretraining, we use the pretrained CLIP-ViT image encoder. Similar to the ViT encoder in the sketch, a candidate scene image $I$ is reshaped to a fixed size ($224 \times 224$) and then spatially sliced into a $16 \times 16$ grid of non-overlapping image patches. Further, these image slices are then reshaped into a sequence of embeddings before passing it to the ViT for further processing. 

\begin{figure}[!t]
    \centering
\includegraphics[width=\columnwidth]{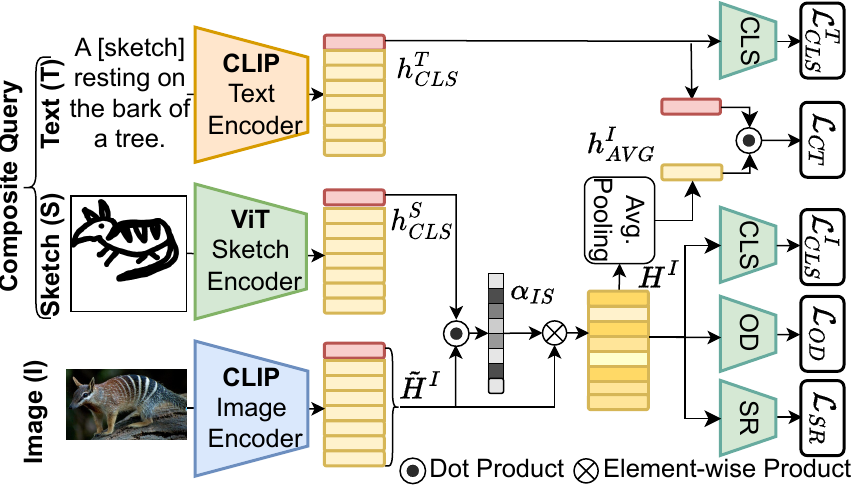}
      \caption{Overview of the proposed method, \model{} for the \data{} problem.}
      \label{fig: method}
\end{figure}

As our problem focuses on queries related to objects in natural scenes, it would be beneficial for our model to focus on the object being queried in the scene image. To enable this, we would like to use the sketch input $S$ to localize or attend to the corresponding object of interest in the image. Specifically, as shown in Figure~\ref{fig: method}, we use the pooled output of the sketch encoder $h_{CLS}^{S}$ to calculate dot-product attention over the output embeddings of the image encoder $\tilde{H}^I$. The obtained values represent attention scores $\alpha_{IS}$ over the spatial regions of the image as well as the $CLS$ token. We obtain weighted values of image embeddings ${H}^{I}$, which are then average pooled to get the final image embedding ${h}^{I}_{AVG}$. Mathematically, $\alpha_{IS} = \text{Softmax}(\tilde{H}^{I} \times h_{CLS}^{S})$, $H^{I} = \alpha_{IS} \odot \tilde{H}^{I}$ and ${h}^{I}_{AVG} = \frac{1}{m}\sum_{i=1}^{m}{H}^{I}_{i}$, where ${h}^{I}_{AVG}$ represents the global average pooled embedding of the image encoder.

\subsection{\model{} Training}

\model{} follows multiple task-specific training objectives.

\subsubsection{Contrastive Training ($\mathcal{L}_{CT}$)}
We adopt a batch-wise contrastive learning strategy akin to CLIP~\cite{radford2021learning} to facilitate image retrieval. Given a batch of $N$ paired $(query, image)$ samples from the train set, we aim to maximize the cosine similarity of the image and query embeddings of the $N$ real pairs in the batch while minimizing the cosine similarity of the embeddings of the $N(N-1)$ incorrect pairings. We use conditional sampling to ensure uniqueness, i.e., a query does not match multiple images and vice versa. Particularly, we use the InfoNCE loss between $h_{CLS}^{T}$ and ${h}^{I}_{AVG}$ to obtain the contrastive loss ($\mathcal{L}_{CT}$) as done in CLIP. Further, as our model utilizes the pretrained CLIP, which lacks joint modeling of text, sketch, and image modalities, we propose three additional training losses to be optimized concomitantly with the contrastive objective.

\subsubsection{Object Classification ($\mathcal{L}_{CLS}^T$ and $\mathcal{L}_{CLS}^I$)}
Given that the \data{} problem focuses on object-specific queries, we propose separately predicting the object name from the text sentence and image inputs. To this end, we train the text and image encoders for the multi-class classification objective to predict the object's class from the $C$ object categories available in the train set. Since the object's label is not mentioned in the text sentence or is always prominent in the image, this objective requires the model to use the contextual information from both modalities to predict the object class. We refer to the classification losses computed using text encodings and the image encodings as $\mathcal{L}_{CLS}^T$ and $\mathcal{L}_{CLS}^I$, respectively.

\subsubsection{Sketch-Guided Object Detection ($\mathcal{L}_{OD}$)}

To aid the localization of the query-relevant object while encoding the image information, inspired by the recent literature in the sketch-guided object detection problem~\cite{tripathi2020sketch,chowdhury2023can,tripathi2023multimodal}, we propose the training objective of sketch-guided localization of the object. Specifically, given the sketch-attended embeddings ${H}^{I}$ from the image encoder, we utilize the embeddings corresponding to the $16 \times 16$ spatial grids. Following the implementation from YOLO~\cite{redmon2016you}, we transform the output embeddings of the ViT network to predict an output of shape $S \times S \times (5B + C)$, where $S \times S$ represents the image grid size, each predicting $B$ bounding boxes, and $C$ class probabilities. We use $S = 7$, $B = 2$, and we have $C = 258$ classes in our train set, so we predict a $7\times7\times268$ output tensor. Finally, we use intersection over union (IoU) to calculate the multipart object detection ($\mathcal{L}_{OD}$) loss as done in~\cite{redmon2016you}.

\subsubsection{Sketch Reconstruction ($\mathcal{L}_{SR}$)}
Similar to the object detection training objective, which facilitates the localization of the relevant objects, we introduce the task of sketch reconstruction using the image features ${H^{I}}$ as illustrated in Figure~\ref{fig: method}. We employ eight blocks of Convolution-BatchNorm-ReLu as done in~\cite{isola2017image} to upsample the information to a reconstructed sketch tensor of size $1\times224\times224$. Further to train the sketch-reconstruction module, we utilize a combination of Binary Cross Entropy loss and the DICE loss~\cite{sudre2017generalised} as $\mathcal{L}_{SR} = \alpha\mathcal{L}_{BCE} + \beta\mathcal{L}_{DICE}$.

Our overall loss is the sum of all five losses $\mathcal{L}_{CT}+\mathcal{L}_{CLS}^T+\mathcal{L}_{CLS}^I+\mathcal{L}_{OD}+\mathcal{L}_{SR}$. 

We measure the distance between the query and the image embedding during retrieval using cosine similarity. We provide the implementation details for \model{} in the Appendix and make code and dataset available at our project page\footnote{\url{https://vl2g.github.io/projects/cstbir}}.

\section{Experiments and Results}
\subsection{Baseline Models}
We compare \model{} extensively with competitive image retrieval baselines.

\noindent\textbf{Sketch-based Image Retrieval (SBIR)}: SBIR is a prominently studied domain in the literature. In our setup, from our composite queries, we only take sketches and drop text to experiment with these baselines. We choose two representative and competitive SBIR methods as our baselines: Doodle2Search~\cite{dey2019doodle} and DeepSBIR~\cite{yu2016sketch}. We also create a Vision Transformer-based SBIR baseline, viz. ViT-based Siamese Network. This network comprises two ImageNet pre-trained ViT-based encoders for sketch and image modalities, trained using the InfoNCE loss~\cite{oord2018infonce}.

\noindent\textbf{Text-based Image Retrieval (TBIR)}: These baselines perform retrieval using only the text part of the query while ignoring the sketch component. We choose the following three modern approaches in this category: VisualBERT~\cite{li2019visualbert}, ViLT~\cite{kim2021vilt}, and CLIP~\cite{radford2021learning}. 

\noindent\textbf{Composite Query-based Image Retrieval}: These baselines perform retrieval using the sketch and text inputs. We compare our proposed method, \model{}, with the following baseline methods: TIRG~\cite{vo2019composing} and TaskFormer~\cite{sangkloy2022sketch}, and a two-stage model. We trained the TIRG model from scratch using our dataset. For Taskformer, we finetuned the publicly available checkpoint using our dataset and our reproduced code for training. We adhered to the hyperparameter configurations outlined in their respective papers for these models. For the two-stage method, in the first stage, we use a ViT trained for sketch classification to get an object name from the sketch. Next, in the second stage, we obtain the full-text query by inserting the predicted object name into the incomplete text and then using pretrained CLIP for image retrieval. 

Finally, we experimented with another baseline, ``two-stage (desc)''. This method's first stage is the same as the ``two-stage''. In the second stage, rather than using the class name, we obtain the full-text query by inserting the \emph{description of the predicted object} into the incomplete text and then using a pretrained CLIP model for image retrieval. The description for each of the 258 object names is chosen randomly from seven different sets of descriptions annotated per object name. Five of these object description sets are obtained automatically, while the other two are manually annotated.

Automated descriptions were generated by using ChatGPT-3.5\footnote{\url{https://chat.openai.com/}} on Mar 14, 2023. We used the following five prompts to obtain five different description sets: (i) ``Describe the following words with visual descriptions  in 4 to 10 words.'' (ii) ``Describe the following words with visual descriptions as a 15-year-old kid in 4 to 10 words.'' (iii) ``Describe the following words with visual descriptions as a 35-year-old in 4 to 10 words.'' (iv) ``Describe the following words with visual descriptions as a 55-year-old in 4 to 10 words.'' (v) ``Describe the following words with visual descriptions as a non-native English speaker in 4 to 10 words.'' The human annotators were asked to write descriptions with 4 to 10 words that included visual attributes without mentioning the object's name.

We use two metrics: Recall@K and Median Rank (MdR). Recall@K is the percentage of times the ground truth image is retrieved within the top $K$ results across all queries in the test set; the higher, the better. Median Rank is the median of the rank of ground truth image in the retrieved set across all queries in the test set; the lower, the better.

\subsection{Results on Test-1K and Test-5K}

\begin{table*}[!t]
    \centering
    \small
      \begin{tabular}{|l|l | r |r |r|r | r|r | r|r | r|r |}
        \hline
        & \multirow{2}{*}{Method}& \multicolumn{2}{r|}{R@10 $\uparrow$} & \multicolumn{2}{r|}{R@20 $\uparrow$} & \multicolumn{2}{r|}{R@50 $\uparrow$} & \multicolumn{2}{r|}{R@100 $\uparrow$} & \multicolumn{2}{c|}{MdR $\downarrow$} \\
        \cline{3-12}
        &&T1K&T5K&T1K&T5K&T1K&T5K&T1K&T5K&T1K&T5K\\
        \hline
        \hline
\multirow{3}{*}{\rotatebox{90}{Sketch}} &Doodle2Search & 14.3&3.6& 24.5&6.7 & 36.2&14.5 & 45.7&24.4 & 129.0&573.5\\
&DeepSBIR & 5.2&1.6 & 8.8&3.0 & 18.9&5.7 & 27.4&9.5 & 258.5&1288.0 \\
&ViT-Siamese & 20.4&5.2 & 34.2&9.9 & 51.0&22.2 & 62.6&34.9 & 48.0&233.0\\
\hline
\multirow{3}{*}{\rotatebox{90}{Text}}&VisualBERT & 23.3&7.6 & 35.9&15.4 & 40.8&27.8 & 54.0&40.2 & 46.0&246.0 \\
&ViLT & 28.1&10.5 & 42.7&16.5 & 60.2&30.1 & 74.3&43.8 & 30.0&163.0 \\
&CLIP & 50.6&24.2 & 63.1&33.7 & 78.8&49.1 & 86.7&62.5 & 10.0&52.0\\
\hline
\multirow{5}{*}{\rotatebox{90}{Sketch+Text}}&TIRG & 31.9&10.4 & 44.2&17.3 & 62.8&31.6 & 73.2&45.4 & 27.5&128.0 \\
&Taskformer & 22.4&9.3 & 35.6&14.8 & 42.3&27.6 & 53.8&38.3 & 48.0&204.0\\
&Two-stage & 67.0&34.8 & 77.4&46.9 & 88.6&64.7 & \textbf{93.7}&\textbf{76.2} & 5.0&24.0\\
&Two-stage (desc) & 60.1&30.5 & 73.7&41.7 & 85.5&59.6 & 91.6&72.0 & 7.0&32.0\\
&\textbf{\model{} (Ours)} & \textbf{73.7}&\textbf{38.7} & \textbf{80.6}&\textbf{50.0} & \textbf{89.4}&\textbf{64.6} & 93.5&74.5 & \textbf{3.0}&\textbf{20.5}\\
        \hline
      \end{tabular}
      \caption{Image retrieval results on \data{} Test-1K (T1K) and Test-5K (T5K). Higher values are preferred for R@K (Recall@K) and lower for MdR (Median Rank).
    }
    \label{table:main-results} 
\end{table*}

Table~\ref{table:main-results} shows our main results on both test sets. Our proposed method, \model{}, outperforms all baseline methods. Multiple sketch+text-based image retrieval models are better than text-based models, which are better than sketch-based image retrieval models. This is mainly because neither the sketches nor the incomplete text can answer the queries accurately. Amongst sketch-based image retrieval models, ViT-based Siamese networks perform the best. Among text-based image retrieval models, CLIP performs the best. \model{} is better than the two-stage model (except for R@100) because the object name may not completely cover the semantics in the sketch and, even worse, may suffer from ambiguous object names (e.g., mouse, bat, star, etc.). 

The two-stage model (desc) is expected to avoid some of the drawbacks of the two-stage model. However, descriptions of object names are often not natural (e.g., a description for ``grass'' is ``green plant used for landscaping and grazing animals'') and are still quite generic. Similarly, consider objects like boat, yacht, ship, and ferry. It is difficult to describe these in a differentiating manner but easy to sketch.
Hence, both the two-stage model and \model{} are better than the two-stage model (desc).

Considering the other sketch+text-based image retrieval models, TIRG~\cite{vo2019composing} and Taskformer~\cite{sangkloy2022sketch}, our proposed model \model{} performs massively better. The poor performance of TIRG is because it does not use any pretraining for text. Also, the image pretraining in TIRG uses ResNet-17~\cite{he2015deep} (trained on ImageNet dataset), which has been shown to lead to poorer image embeddings compared to CLIP~\cite{radford2021learning}. For Taskformer, we finetuned the publicly available checkpoint using ouris because the initial checkpoint has been trained on a dataset where the (a) images in the collection are focused object images, unlike scene images in our dataset, (b) sketches are elaborate and not crudely drawn, and (c) text is self-contained and not incomplete. In other words, the samples on our dataset, \data{}, are more challenging (closer to practical settings) compared to data used to train Taskformer. We also experimented with training the Taskformer model from scratch but did not see any improvements. Finally, Taskformer does not use sketch reconstruction and object detection losses, which cater to the object-centric nature of our dataset, as shown in Table~\ref{tab:ablation}.

\begin{figure*}[!t]
    \centering
\includegraphics[width=0.90\textwidth]{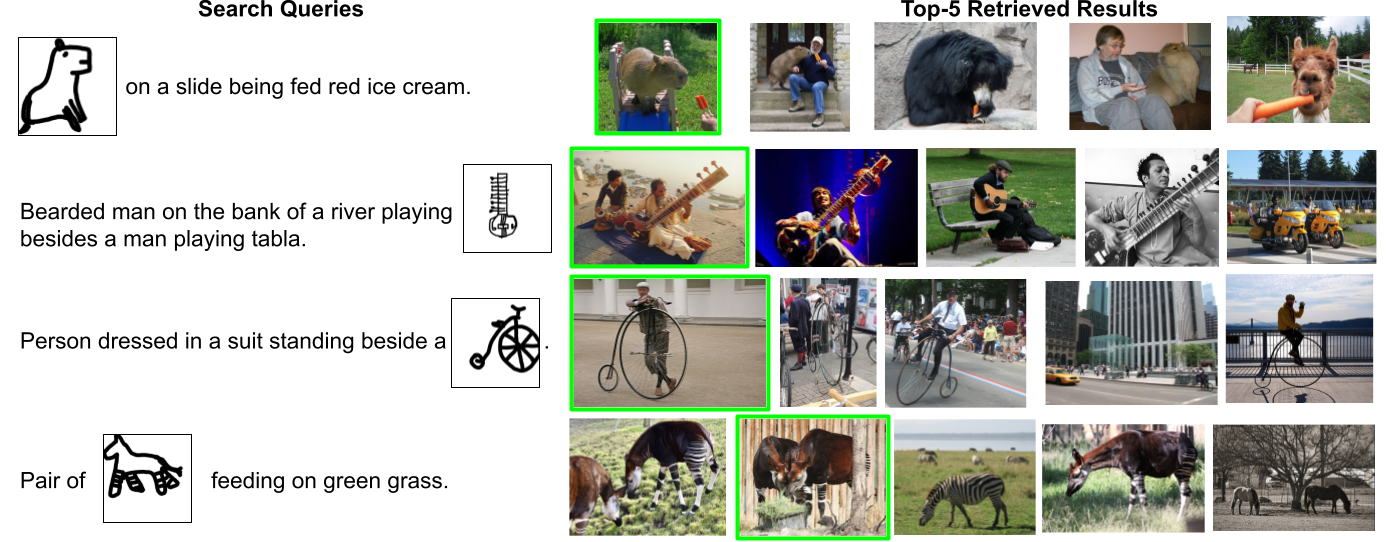}
      \caption{Qualitative results of our \model{}. We show top-5 retrieved results for the multimodal (sketch+text) queries shown in left most column. From top to bottom, the sketch are for capybara, sitar, penny-farthing, and okapi. The ground truth image is shown with a green frame. (Best viewed in color).}
      \label{fig:stnet-qualitative-results}
\end{figure*}

\subsection{Ablation Study}
Our overall \model{} model consists of several components. To understand the importance of each component, we perform several ablations as shown in Table~\ref{tab:ablation}. 

\setlength{\tabcolsep}{2pt}
\begin{table}[!t]
    \centering
    \small
      \begin{tabular}{|c|c|p{0.7in}|r|r|r|r|r|r|}
        \hline
        M&Query&Objective&R@10&R@20&R@50&R@100&MdR\\
        \hline
        \hline
1&S&\loss{CT}&20.2&33.7&50.9&62.9&50.5 \\
2&T&\loss{CT}&50.6&63.1&78.8&86.7&10.0 \\
3&T+S&\loss{CT}&68.4&77.2&85.6&89.8&5.0 \\
4&T+S&\loss{CT} + \loss{OD} + \loss{SR}&69.4&80.4&85.6&90.4&5.0 \\
5&T+S&\loss{CT} + $\mathcal{L}_{CLS}$ + \loss{SR}&70.4&79.6&86.2&91.1&5.0 \\
6&T+S&\loss{CT} + $\mathcal{L}_{CLS}$ + \loss{OD}&71.2&79.0&87.0&93.0&4.0 \\
7&T+S&(6) + \loss{SR}&\textbf{73.7}&\textbf{80.6}&\textbf{89.4}&\textbf{93.5}&\textbf{3.0} \\
        \hline
      \end{tabular}
    \caption{Ablation study for \model{} on Test-1K set based on query modalities and training objectives. Models (M) 1 and 2 are text-only (T) and sketch-only (S) query-based methods, resp. Models 3-6 denote objective-based ablations. Model 7 is our final model. (\loss{CT}: contrastive loss, \loss{CLS}: classification loss, \loss{OD}: object-detection loss, and \loss{SR}: sketch-reconstruction loss). Higher values are preferred for recall and lower ones for MdR. $\mathcal{L}_{CLS}$=$\mathcal{L}^T_{CLS}$+$\mathcal{L}^I_{CLS}$.}
\label{tab:ablation}
\end{table}
\setlength{\tabcolsep}{3pt}

We start with just the contrastive loss ($\mathcal{L}_{CT}$) computed using sketch modality alone (Model 1). Model 2, which is trained with just $\mathcal{L}_{CT}$ computed using only text modality, performs better. This broadly indicates that the information in text is higher than in sketch, which makes sense since our sketches are quite rough. Using text and sketch for contrastive loss computation (Model 3) leads to further improvements. Note that we do not perform dot-product attention between sketch and image in Model 1; rather, we employ contrastive learning between their encoders.
Our full proposed model, \model{} (Model 7), consists of all the loss functions: contrastive loss ($\mathcal{L}_{CT}$), object classification loss using text encodings ($\mathcal{L}_{CLS}^T$), object classification loss using image encodings ($\mathcal{L}_{CLS}^I$), sketch-guided object detection loss ($\mathcal{L}_{OD}$) and sketch reconstruction loss ($\mathcal{L}_{SR}$). 
Models 4, 5 and 6 are trained by removing classification ($\mathcal{L}_{CLS}^T$+$\mathcal{L}_{CLS}^I$) loss, object detection loss ($\mathcal{L}_{OD}$) and sketch reconstruction loss ($\mathcal{L}_{SR}$) respectively from the overall \model{} model. Broadly, removing any of the three losses leads to degradation in performance across all metrics compared to the full \model{} model (Model 7). The degradation worsens when the $\mathcal{L}_{CLS}$ is removed (Model 4). 

\subsection{Results on Open-Category Test Set}

In a real-world scenario, the objects in queries may be uncommon or entirely unfamiliar. Considering that the Visual Genome focuses solely on common objects, we curate an Open-Category test set featuring 70 novel object categories under nine overarching classes. Among these, 50 are rare objects that are challenging to name but simple to illustrate, examples being Numbat, Mangosteen, Feijoa, Draw Knife, and Gibraltar Campion. These objects and their corresponding sketches are entirely unseen in the training set. The classes are mentioned in the Appendix. This set includes 750 composite queries and 1K gallery images.

\begin{table}[!t]
    \centering
    \small
      \begin{tabular}{|l |r| r| r| r| r|}
        \hline
Method&R@10$\uparrow$&R@25$\uparrow$&R@50$\uparrow$&R@100$\uparrow$&MdR$\downarrow$\\
\hline
\hline
ViT-Siamese&6.3&8.6&14.5&23.0&241.0\\
\hline
CLIP&21.6&30.6&39.4&47.6&71.0\\
\hline
Two-Stage&29.0&38.2&48.8&54.8&63.0\\
\hline
\model{} (Ours)&\textbf{37.2}&\textbf{45.3}&\textbf{62.3}&\textbf{71.7}&\textbf{27.5}\\
\hline
      \end{tabular}
      \caption{Performance of image retrieval for object classes that are unseen during training. This measures the ability of the baselines to generalize concepts outside of the training domain. We evaluate this on an Open-Category test set of 750 samples containing 70 unseen object classes.}
      \label{table:unseen-classes}
\end{table}

Table~\ref{table:unseen-classes} showcases results for this experiment, comparing \model{} to the top sketch-only (ViT-Siamese), text-only (CLIP), and sketch+text (Two-Stage) baselines. Although \model{} is naturally extensible to novel object categories, the two-stage model requires a pre-defined universe of possible objects to select from. Hence, we first create a set of possible object categories for the two-stage model by augmenting the train set with the 70 additional test categories. ViT can't extend to new classes during testing, so we use zero-shot CLIP for sketch classification in the two-stage baseline. From Table~\ref{table:unseen-classes}, we observe that (i) the Open-Category setting is difficult as expected. (ii) Since \model{} encodes generic visual semantics from sketches, it is more robust to this complex setting than all the baselines.

\begin{table}[!t]
    \centering
    \small
      \begin{tabular}{|l |r| r| r| r| r|}
        \hline
        Method&R@10$\uparrow$&R@25$\uparrow$&R@50$\uparrow$&R@100$\uparrow$&MdR$\downarrow$\\
\hline
\hline
ViT-Siamese&41.5&50.3&58.6&63.1&17.0\\
\hline
CLIP&50.6&63.1&78.8&86.7&10.0\\
\hline
Two-Stage&61.4&72.5&82.8&89.3&7.0\\
\hline
\model{} (Ours)&\textbf{70.3}&\textbf{81.8}&\textbf{90.7}&\textbf{95.6}&\textbf{3.0}\\
\hline
      \end{tabular}
      \caption{Performance of image retrieval for examples with instance-level sketches. This measures the ability of the baselines to generalize to rich sketches with pose, size, and shape information. We evaluate this on Test-1K (where rich ones have replaced crude sketches).}
      \label{table:instance-level}
\end{table}

\subsection{Performance with Instance-Level Sketches}
We have primarily focused on crude sketches. How does \model{} fare with detailed instance-level sketches—those with pose, size, and shape details? Such sketches require the retrieved image to have a matching object instance. For this experiment, we generate rich synthetic sketches automatically for each image in the Train and Test-1K datasets using the method proposed in~\cite{li2019photo}. We obtain sketches only for the part of the image covered by the relevant object box. Table~\ref{table:instance-level} shows that \model{} outperforms all baselines on this complex setting as well. As the sketch becomes more expressive, the two-stage model, converting the sketch to a category name, loses nuanced details, widening its gap with \model{}. More details are in the Appendix.

\subsection{Qualitative Analysis} 
We show a few retrieval results from the \data{} dataset for our model \model{} in Figure~\ref{fig:stnet-qualitative-results}. Our model correctly retrieves the ground truth image associated with each composite query and ranks several relevant images in the top results. We observe that it can even reason about certain complex visual attributes associated with the queried object (e.g., ``okapi'' with striped legs). We provide more analyses in the appendix. 

\section{Conclusion}
We proposed the novel problem of multimodal query-based retrieval on a collection of natural scene images where the query consists of an incomplete text and an accompanying rough sketch. Towards this task, we contributed a novel dataset, \data{}, containing $\sim$2M queries and $\sim$103K natural scene images. Further, we also proposed a novel model, \model{}, which is trained on losses specially designed for the \data{} problem: contrastive loss, object classification loss, sketch-guided object detection loss, and sketch reconstruction loss. \model{} outperforms existing strong baselines by significant margins. \data{} could be essential in multiple real-world settings. For example, searching for a product in digital catalogs given its rough sketch and a short description. It can also aid in the search for missing people, given their prominent features with accompanying descriptions from a repository of crowd photos taken by surveillance cameras. 

\section*{Acknowledgements} This work is supported by the Startup Research Grant from the Science and Engineering Research Board (SERB), Department of Science and Technology, Government of India (Grant No: SRG/2021/001948).

\bibliography{refs}
\end{document}